\documentclass[manuscript,screen,authorversion,nonacm]{acmart}

\AtBeginDocument{%
  \providecommand\BibTeX{{%
    \normalfont B\kern-0.5em{\scshape i\kern-0.25em b}\kern-0.8em\TeX}}}

\acmBooktitle{FGVRreID2021: ACM Special Issue on Fine-Grained Visual Recognition and re-IDentification}

\newcommand{\error}[1]{\textcolor{blue}{#1}}
\usepackage{subfigure}
\usepackage{graphicx}
\usepackage{amsmath}
\usepackage{float}
\usepackage{wrapfig}
\usepackage{xcolor}
\usepackage{caption}

\begin{document}

\received[accepted]{7th September 2021}

\setcopyright{acmcopyright}
\acmJournal{TOMM}
\acmYear{2021} \acmVolume{1} \acmNumber{1} \acmArticle{1} \acmMonth{1} \acmPrice{15.00}\acmDOI{10.1145/3485473}

\title{Fine-Grained Adversarial Semi-supervised Learning}

\author{Daniele Mugnai}
\affiliation{%
  \institution{University of Florence}
  \streetaddress{Via S. Marta 3}
  \city{Florence}
  \country{Italy}}
\email{daniele.mugnai@unifi.it}

\author{Federico Pernici}
\affiliation{%
  \institution{University of Florence}
  \streetaddress{Via S. Marta 3}
  \city{Florence}  
  \country{Italy}
  \postcode{50139}}
\email{federico.pernici@unifi.it}

\author{Francesco Turchini}
\affiliation{%
  \institution{University of Florence}
  \streetaddress{Via S. Marta 3}
  \city{Florence} 
  \country{Italy}}
\email{francesco.turchini@unifi.it}

\author{Alberto Del Bimbo}
\affiliation{%
  \institution{University of Florence}
  \city{Florence}  
    \country{Italy}}
\email{alberto.delbimbo@unifi.it}

\renewcommand{\shortauthors}{Mugnai and Pernici, et al.}

\begin{abstract}
In this paper we exploit Semi-Supervised Learning (SSL) to increase the amount of training data to improve the performance of Fine-Grained Visual Categorization (FGVC). 
This problem has not been investigated in the past in spite of prohibitive annotation costs that FGVC requires. Our approach leverages unlabeled data with an adversarial optimization strategy in which the internal features representation is obtained with a second-order pooling model. 
This combination allows to back-propagate the information of the parts, represented by second-order pooling, onto unlabeled data in an adversarial training setting. 
We demonstrate the effectiveness of the combined use by conducting experiments on six state-of-the-art fine-grained datasets, which include Aircrafts, Stanford Cars,  CUB-200-2011, Oxford Flowers, Stanford Dogs, and the recent Semi-Supervised iNaturalist-Aves. Experimental results clearly show that our proposed method has better performance than the only previous approach that examined this problem; it also obtained higher classification accuracy with respect to the supervised learning methods with which we compared.
\end{abstract}



\maketitle

\section{Introduction}
Fine-Grained Visual Categorization (FGVC) lies in the continuum between categorization (i.e object classification) and identification (i.e. instance recognition). 
FGVC is quite subtle and therefore difficult to address with general-purpose object classification methods based on Deep Neural Networks (DNNs) \cite{krizhevsky2012imagenet}. 
FGVC is much more challenging than traditional classification tasks due to the inherently subtle intra-class object variability amongst sub-categories.
Distinguishing between a cat and a giraffe is easy (i.e. large variability) while, in distinguishing fine-grained classes, typically only a few key features matter as in species of birds \cite{wah2011caltech}, dogs \cite{KhoslaYaoJayadevaprakashFeiFei_FGVC2011}, flowers \cite{Nilsback08} or manufacturers and models of cars \cite{krause20133d} and aircrafts \cite{maji2013fine}.

\begin{figure}
    \includegraphics[width=0.6\columnwidth]{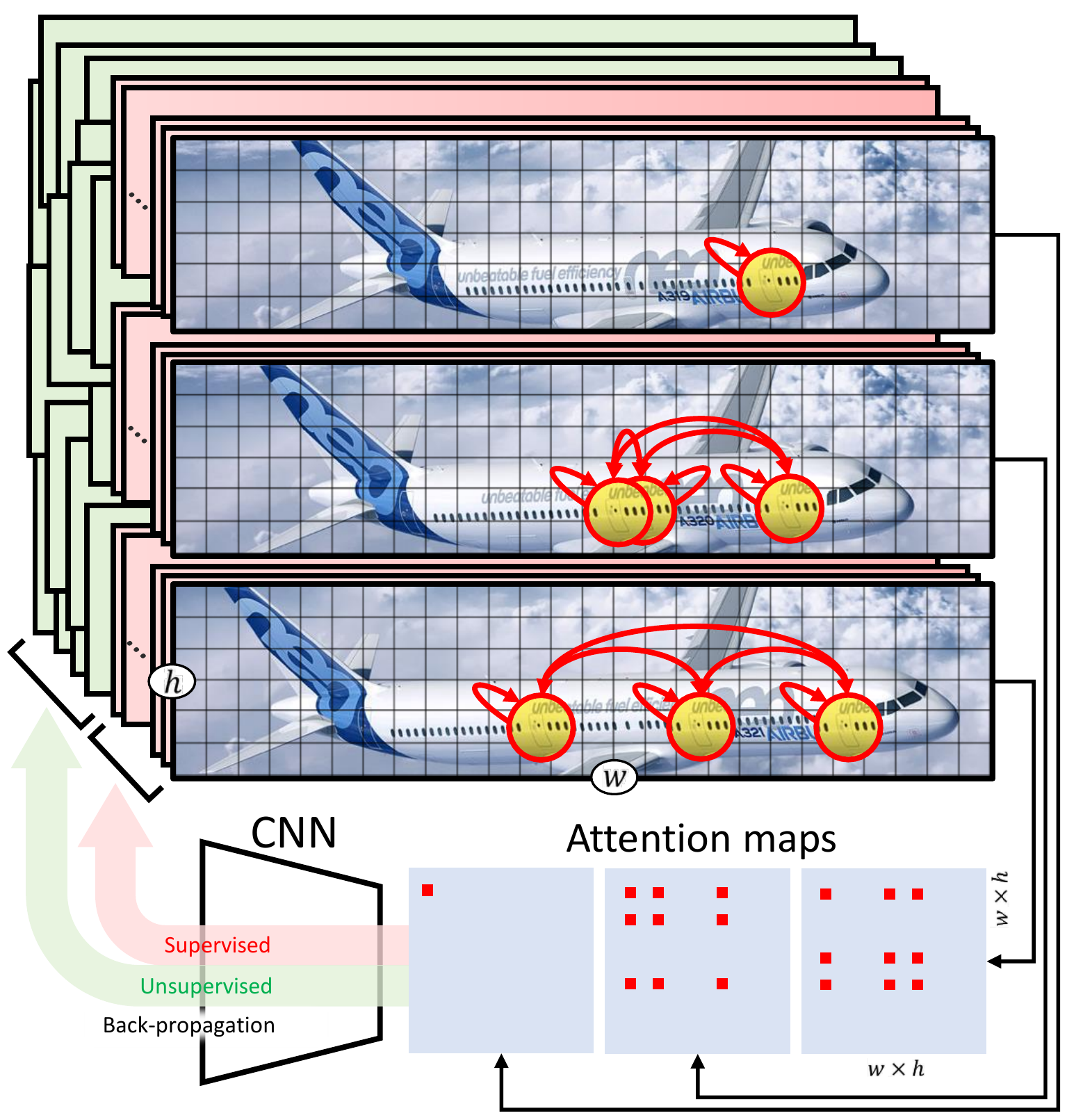}
    \caption{Illustration of the effect of second-order pooling in Semi-Supervised Fine-Grained Visual Categorization (SSL-FGVC). We show images from three different classes of Airbus aircraft models: A319 (\emph{top}), A320 (\emph{middle}) and A321 (\emph{bottom}). They mainly differ by the number of doors and their position along the fuselage (circles). %
    We propose to take advantage of the long-range attention based part-to-part relationships exploited by second-order pooling and back-propagate this information onto unlabeled data to perform unsupervised structure discovery.
    }
\label{fig:intro}
\end{figure}

The task becomes significantly more difficult in domains where data is not readily available (e.g., medical images) or domains for which training data is scarce \cite{wei2019deep}.
It is likely that techniques used for representation learning like semi/self-supervised or unsupervised learning that are currently used for visual recognition are not sufficient to significantly improve FGVC.
In addition to this, obtaining training data for fine-grained images is prohibitively expensive, as expert knowledge is typically required \cite{deng2013fine}. 
In view of these issues, we propose a learning method focusing on the FGVC problem in which labeled data is limited and unlabeled is available.

Recent top performing supervised learning methods have substantially shown that the most successful strategy to FGVC is obtained by identifying, either \emph{explicitly} or \emph{implicitly}, the object parts \cite{korsch2020end,zhang2020three,ngiam2018domain,dosovitskiy2020image}. The central underlying assumption is that fine-grained information resides within the parts. Many approaches particularly focus on explicitly localizing relevant regions in an image. 
This is typically achieved by leveraging the extra annotations of bounding box and part annotations (some known datasets provide ground-truth part annotations \cite{wah2011caltech, berg2014birdsnap}) to localize regions that provide the most discriminative information.  However, in addition to class labeling, the extra human annotations regions are not only difficult to obtain and prohibitively expensive, but can often be error-prone resulting in performance degradation \cite{zheng2017learning}.
Methods for unsupervised part detection and mining have been developed \cite{ge2019weakly,korsch2019classification,zhang2019learning,zhang2019unsupervised}, 
however, these methods pose various challenges, such as missing parts due to occlusions and parts not providing discriminative information. According to this, it remains controversial whether unsupervised detected parts are fully beneficial.

Despite these efforts addressing different aspects regarding extra parts annotation, we are aware of only one published work in the literature addressing FGVC in a Semi-Supervised Learning setting at the image label level \cite{nartey2019semi}. Despite not being investigated, this topic is getting increasing attention and support. 
In confirmation of this, a dataset\footnote{The Semi-Supervised iNaturalist-Aves Dataset: \url{https://github.com/cvl-umass/semi-inat-2020}} for this specific problem has been recently released for the challenge part of the FGVC7
workshop \cite{su2021semisupervised} held in conjunction with CVPR2020. 
The dataset is intended to set out some of the difficulties faced in a practical environment. 
The competition panel reported that all teams applied the pseudo-label SSL method \cite{lee2013pseudo} and that the state-of-the-art Deep-SSL methods \cite{zhai2019s4l,tarvainen2017mean, berthelot2019mixmatch, sohn2020fixmatch} provides similar performance but are  computationally more expensive.

According to this, 
we propose an approach that addresses the problem in a \textit{complementary} way in both the SSL setting and the part-based assumption of FGVC. We adopt an adversarial optimization strategy that alternately maximizes the conditional entropy of unlabeled data with respect to the classifier and minimizes it with respect to a second-order feature encoder. 
This combination allows to \textit{back-propagate} the information of the \textit{parts} captured by the second-order pooling model onto \textit{unlabeled data} in an adversarial training setting as illustrated in Fig.~\ref{fig:intro}. 
The strategy extends the works in  \cite{ganin2015unsupervised,ganin2016domain}, originally proposed for Domain Adaptation, to the specific Semi-Supervised Learning setting. 
To the best of our knowledge, this is the first approach to leverage adversarial optimization in the specific case of SSL, and we are not aware of previous works that combine SSL with specific strategies exploiting the information of object parts. 
Although the recent work \citep{nartey2019semi} applies SSL to FGVC datasets, the information of the parts is not explicitly taken into account during the semi-supervised learning process.
We have reported a few preliminary baseline evaluations of our investigation in
\cite{mugnai2021soft}.

The main benefit of our adversarial optimization with respect to pseudo-label based methods \cite{lee2013pseudo,nartey2019semi}, is that the model can correct its own errors without incurring in wrong classifications that rapidly intensify resulting in confident but erroneous pseudo-labels on the unlabeled data. This soft label assignment may also have implications in those Deep Learning contexts in which data comes from unlabeled video streams \cite{pernici2017unsupervised,pernici2018memory,pernici2020self}.

We empirically demonstrate the superiority of our method over many baselines and show the method is safe \cite{wang2013safety} and \cite{li2019safe}.
\section{Related Work}
\subsection{Fine Grained Visual Categorization} 
Recently the accuracy of FGVC benchmarks has been substantially improved using Deep Learning. Despite the complexity of the problem \cite{anderson2020facing}, several methods emerged. These methods can be broadly classified in two main categories: global and localization-based \cite{wei2019deep}. Global methods use the input image as a whole and employ different strategies for pre-training \cite{cui2018large}, augmentation \cite{touvron2019fixing,krause2016unreasonable}, or pooling \cite{lin2015bilinear, simon2017generalized, simon2018whole, zheng2019learning} to exploit the parts. In contrast, localization-based methods approaches apply sophisticated detection techniques in order to determine the regions of the parts \cite{zhang2014part, zhang2016spda, ge2019weakly,korsch2019classification,yang2018learning,zhang2019learning,zhang2019unsupervised, korsch2020end, xiao2014application}. %
The two categories have shown comparable results.

Recent state-of-the-art methods follow different strategies. The work \cite{korsch2020end} proposes a layer based on Fisher Vector encoding for aggregating part features, \cite{zhang2020three} proposes a multi-scale object and part attention approach { and the work \cite{yang2018learning} uses an unsupervised learning scheme  to localize informative regions}. The work in \cite{ngiam2018domain} exploits both large external labeled datasets (the JFT \cite{sun2017revisiting} private dataset with 300M images) and large architectures (the AmoebaNet architecture \cite{real2019regularized} with 550M parameters). It shows that fine-grained labels of JFT achieve better transfer learning for FGVC datasets.
In \cite{dosovitskiy2020image} the transformer model \cite{vaswani2017attention} evaluated in FGVC benchmarks achieves significant performance improvements.
Despite the promising results these approaches strongly rely on supervision, large model architecture or sophisticated part-based models.

The most representative method among the end-to-end feature encoding category, not relying on external part detector or complex augmentation schemes, is Bilinear-CNN \cite{lin2015bilinear}. Bilinear-CNN  represents an image as a pooled outer product that captures pairwise correlations between the convolutional feature channels and can model part-feature interactions. The method enhances the mid-level learning capability achieving clear performance improvement on FGVC.
A similar approach has been concurrently proposed in \cite{ionescu2015matrix}. These methods have been extended and improved in several different aspects. 
In \cite{BMVC2017_117}, the use of Newton-Schulz iteration \cite{higham2008functions} for efficient and accurate back-propagation is firstly proposed and in \cite{li2018towards} strategy is further improved. We follow this approach
to implicitly improve the discriminativity of the feature representation in unsupervised structure recovery.

FGVC is also interesting because it is related to the problem of learning features suitable for visual search problems such as image retrieval \cite{chen2021deep}, face recognition \cite{WANG2021215} and person re-identification \citep{ye2021deep}. Moreover, fine grained datasets are shown to be less susceptible to catastrophic forgetting in class-incremental learning \cite{masana2020class} and preliminary evidence in \cite{ChenLWTBL20} shows encouraging results in fine-grained continual image retrieval. For the continual re-identification problem, \cite{Pu_2021_CVPR} shows that performance does not decrease during learning from multiple datasets. This evidence opens up also to recent methodologies for learning the features so-called compatible \cite{shen2020towards}, based on the concept of fixed classifiers \cite{Pernicicvprw19,pernici2020icpr,perniciTNNLS2021}, in which the re-indexing of the gallery is no longer necessary when upgrading the feature representation. 

\subsection{Semi-Supervised Learning} 
There is a wide literature on SSL techniques that we do not discuss here. Comprehensive
overviews are provided in \cite{zhu2003semi,chapelle2010semi} and more recently in \cite{van2020survey, ouali2020overview} for Deep SSL methods.
SSL algorithms are typically 
\cite{oliver2018realistic, miyato2018virtual, berthelot2019mixmatch, athiwaratkun2018there, liu2019deep, yalniz2019billionscale}
evaluated on small-scale datasets such as CIFAR-10 \cite{krizhevsky2009learning} and SVHN \cite{netzer2011reading}. We are aware of very few examples in the literature where SSL algorithms are evaluated on more challenging datasets such as the FGVC ones. %
To our knowledge, Mean Teacher \cite{tarvainen2017mean} currently holds the state-of-the-art result on ILSVRC-2012 \cite{russakovsky2015imagenet} when using only 10\% of the labels and \cite{nartey2019semi} holds the state state-of-the-art result on FGVC datasets which has been improved in this paper. Recent concurrent works \cite{henaff2019data, xie2019unsupervised} present competitive results on ILSVRC-2012 \cite{russakovsky2015imagenet}.

A well-known strategy in semi-supervised learning with neural networks follows the concept of pseudo-label \cite{lee2013pseudo}. In pseudo-labeling method proposed network predictions are used as labels
for unlabeled data: in particular a pseudo-label is assigned to the class which has the maximum predicted probability. This class is considered as if it were the true class. As discussed in \cite{lee2013pseudo} this method is equivalent to entropy regularization \cite{grandvalet2005semi} and the conditional entropy of the class probabilities can be used for a measure of class overlap. By minimizing the entropy for unlabeled data, the overlap of class probability distribution can be reduced. FixMatch \cite{sohn2020fixmatch}, one of the best performing methods, combines pseudo-labeling and regularization applying a strong augmentation on unlabeled data. Another recent well performing method is \citep{cascantebonilla2020curriculum} which
generates pseudo-labels trained under curriculum labeling.

The main downside of such methods is that the model is unable to correct its own mistakes
\cite{zhai2019s4l, tarvainen2017mean, berthelot2019mixmatch, sohn2020fixmatch}.
We address this issue with a complementary approach based on the so called Gradient Reversal Layer (GRL) firstly introduced in \cite{ganin2015unsupervised} and in \cite{ganin2016domain} for the Unsupervised Domain Adaption (UDA) and the Semi-Supervised Domain Adaptation problem (SSDA) respectively.
SSL shares similarities with the related topic of Semi-Supervised Domain adaptation (SSDA), in which the test and train samples come from two different distributions \cite{daume2010frustratingly, kumar2010co, donahue2013semi, yao2015semi, saito2019semi}. However, standard evaluation of SSL algorithms typically does not consider this case. Moreover, SSDA requires annotated data in the target distribution, and therefore it is not well suited to fine-grained data due to the cost of expert annotation.
In \cite{ganin2016domain} the authors left the extension and the verification of their work to the case of Semi-Supervised setting for future work. In this paper we extend the GRL to the SSL problem and evaluate it with FGVC datasets.

GRL has been used for other tasks mostly inspired by \textit{Domain Adaptation}. Some works applied GRL for the task of object detection \citep{he2019multi} and \citep{li2018mixed} to transfer the knowledge of the strong categories to the weak categories, for the task of image segmentation \citep{javanmardi2018domain}, speech synthesis \citep{zhang2019learning}, speech recognition \citep{shinohara2016adversarial} and 3D object reconstruction \citep{kato2019learning}. 

\subsection{Semi-Supervised FGVC.} 
The method \citep{nartey2019semi} uses pseudo-labels \citep{lee2013pseudo} to handle the unlabeled samples and proposes a novel strategy to select the unlabeled samples that have the highest confidence. Differently from the work presented in this paper, the work \citep{nartey2019semi} uses CNN models with a linear classifier and therefore the learned fine grained structure is not explicitly based on the parts of the objects.

Only very recently, the Semi-Supervised iNaturalist-Aves dataset has been released \cite{su2021semisupervised}. The dataset is designed to expose some of the challenges encountered in a realistic setting, such as the ﬁne-grained similarity between classes, significant class imbalance, and domain mismatch between the labeled and unlabeled data. The competition panel reported that all teams applied the pseudo label method \cite{lee2013pseudo} and that the state-of-the-art method \cite{zhai2019s4l} provides similar performance but is computationally more expensive. It is further reported that other recent state-of-the-art methods \cite{tarvainen2017mean, berthelot2019mixmatch, sohn2020fixmatch} do not substantially improve the performance. These results have been confirmed very recently in the evaluation performed in \cite{su2021realistic}. The  experimental evaluation further shows the limitations of existing methods in taking advantage of out-of-class unlabeled data.

The only approach we are aware of using the Semi-Supervised iNaturalist-Aves is \cite{cui2020semi}, in which an ensemble of 8 deep CNN models is learned according to the pseudo label method \cite{lee2013pseudo}, and the WS-DAN \cite{hu2019see}. Many several strategies are further included to improve the final performance of more than 20\% points of accuracy. Out-of-class data is also used to cluster classes labels that are used to pre-train the models of the ensemble.

\section{Contributions}
\begin{itemize}
    \item 
    We firstly propose Semi-Supervised Learning to take advantage of objects parts in FGVC to reduce the prohibitive cost of expert annotation of data.
    \item We demonstrate the effectiveness of our method in comparison with state-of-the-art and carefully tuned approaches. In particular, we improve the state-of-the-art in the Semi-Supervised iNaturalist-Aves dataset from 57.4\% to 65.4\%.
    \item %
    We extend and verify the Gradient Reversal Layer (GRL), originally proposed in \cite{ganin2016domain} 
    for Domain Adaptation, to the Semi-Supervised Learning setting.
\end{itemize}
\section{Problem Formulation}

In Semi-Supervised Learning, in addition to unlabeled data, the learning algorithm is provided with some supervision information, but not necessarily for all examples. In this case, the data is divided in two parts: the set for which labels are available $\mathcal{D}_{l}=\left\{\left(\mathbf{x}_{i}^{l}, y_{i}^{l}\right)\right\}_{i=1}^{n_{l}}$ and the set for which the labels are unknown $\mathcal{D}_{u}=\left\{\left(\mathbf{x}_{i}^{u} \right)\right\}_{i=1}^{n_{u}}$, where $n_l$ and $n_u$ are the number of examples of the labeled and unlabeled datasets, respectively.
It is typically assumed that 
$\mathbf{X} \times Y$ is drawn from an unknown joint probability distribution $p(\mathbf{X},Y)$ and that we observe it through the finite training sample $\mathcal{D}_{l}$.
The main goal is to leverage the unlabeled data $\mathcal{D}_{u}$ to learn a Deep Neural Network model $f_\theta$, with trainable parameters $\theta$, that is more accurate than using the only $\mathcal{D}_{l}$. The data $\mathcal{D}_{u}$ provide additional information about the structure of the data distribution $p(\mathbf{X})$ to better learn the internal feature representation of $f_\theta$. The dependency between $p(\mathbf{X})$ and $p(Y | \mathbf{X})$ is typically established according to the \textit{cluster assumption}, (i.e. data points in the same cluster of $p(\mathbf{X})$ have the same label $Y$); and \textit{low-density separation}, (i.e. class boundaries of $p(Y |\mathbf{X})$ should lie in an area where $p(\mathbf{X})$ is small) \cite{chapelle2010semi}. 
Due to the low variability of classes in FGVC, these assumptions are quite hard to meet in practice.

\begin{figure}[t]
    \vspace{1.1cm}
    \includegraphics[width=0.50\columnwidth]{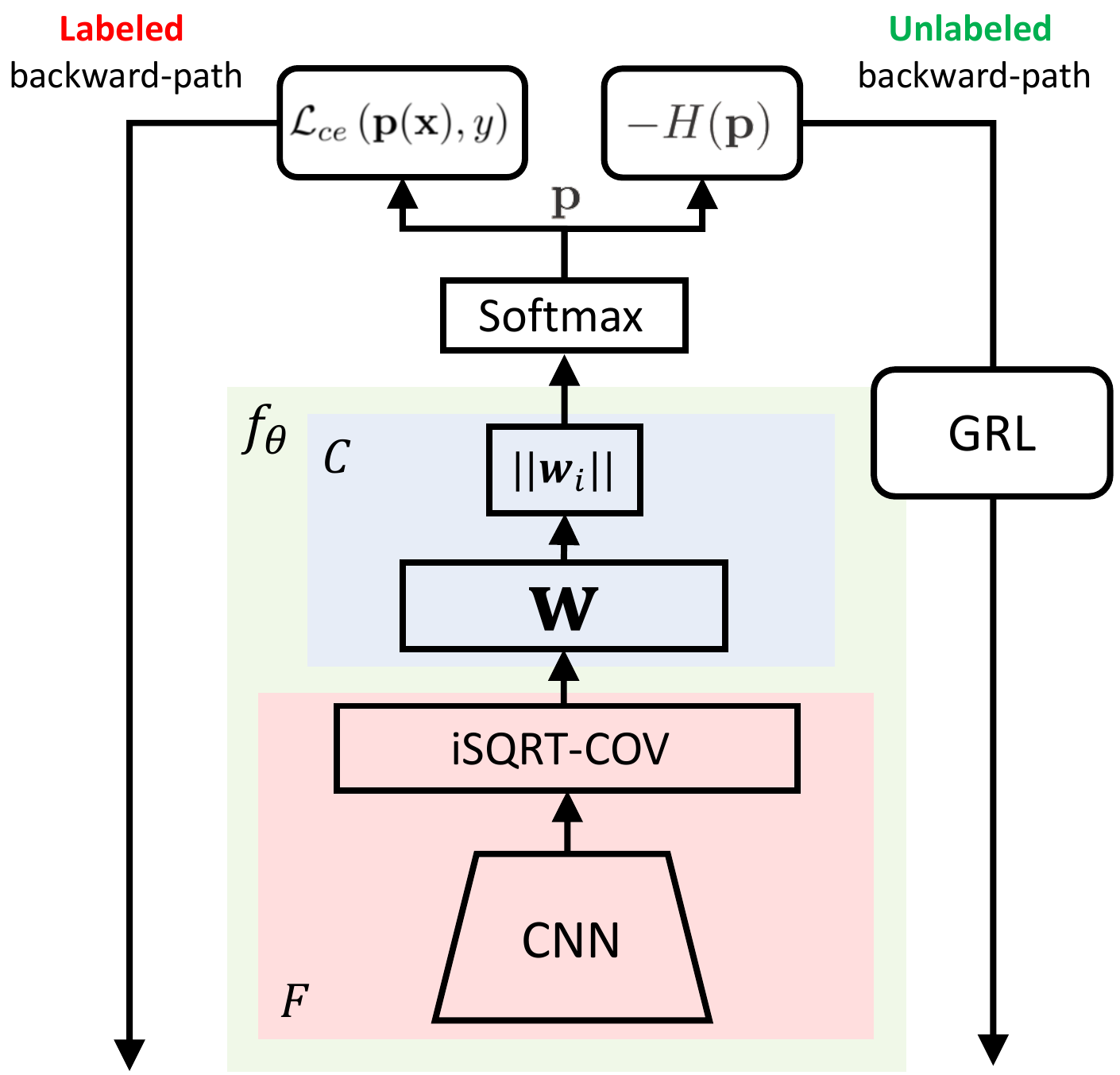}
    \caption{
    An overview of the proposed model architecture. The inputs to the network are labeled and unlabeled examples. The model $f_\theta$ (light green) consists of the second-order pooling (iSQRT-COV) \cite{MPN-COV_PAMI} feature extractor $F$ (light red) and the classifier $C$ having weight vectors $\mathbf{w}_i$ (light blue). $C$ is trained to maximize entropy on unlabeled target whereas $F$ is trained to minimize it. To achieve the adversarial learning, the sign of gradients for entropy loss on unlabeled target examples is flipped by a gradient reversal layer (GRL) \cite{ganin2015unsupervised}. According to this, labeled and unlabeled back-propagation follows two distinct paths.
    }
    \Description{Illustration of the model architecture composed by a feature extractor that is followed by a normalize linear classifier, whose output is fed into a softmax layer that produce a probability distribution. The probability distribution is used on two different path for optimization depending on data whether labeled or not.}
    \label{fig:architecture}
\end{figure}

\subsection{Method Overview}
Our base model architecture $f_\theta$ for FGVC consists of a special feature extractor $F$, based on second-order pooling, and a classifier $C$ in which weights $\mathbf{W}$ are normalized to exploit the approximated cosine distance criterion between the classifier prototypes and the features.
According to this, in order to classify examples correctly, the normalized direction of a weight vector has to be representative of the features of the corresponding class in term of an angular distance. In this respect, the weight vectors can be regarded as angular estimated prototypes for each class. 

Angular classifier prototypes are learned taking into account unlabeled data by exploiting an adversarial entropy optimization. Unlabeled data follow a specific data path for back-propagation that allows to attract the prototypes towards them. 
As the feature representation {takes in account} the long-range part-to-part relationships, the {information} of the parts can be indirectly better back-propagated towards the unlabeled data. 
The architecture of our method is shown in Fig.~\ref{fig:architecture} and will be detailed in the next Subsection.

\subsection{Back-propagation of the Parts onto Unlabeled Data} 
\label{subsection:Second-order pooling SSL Adversarial Optimization}

The goal is to obtain representative classifier prototypes 
$\mathbf{W} = [\mathbf{w}_1, \mathbf{w}_2, \dots, \mathbf{w}_{K}]$ where ${K}$ is the number of classes for labeled and unlabeled data and to minimize the distance between the prototypes and the unlabeled examples.
We approach the problem in mainly three different fronts: 1) reduce intra-class distance (next paragraph), 2) improve feature discriminativity exploiting part-to-part relationships (Subsection~\ref{subsec:Second-Order Pooling Feature Extraction}) 3)  handling unlabeled data distance between prototypes and features (Subsection~\ref{subsec:Handling Unlabeled Data with Adversarial Entropy Optimization}). 

\begin{figure*}[ht]
    \centering
    \includegraphics[width=\textwidth]{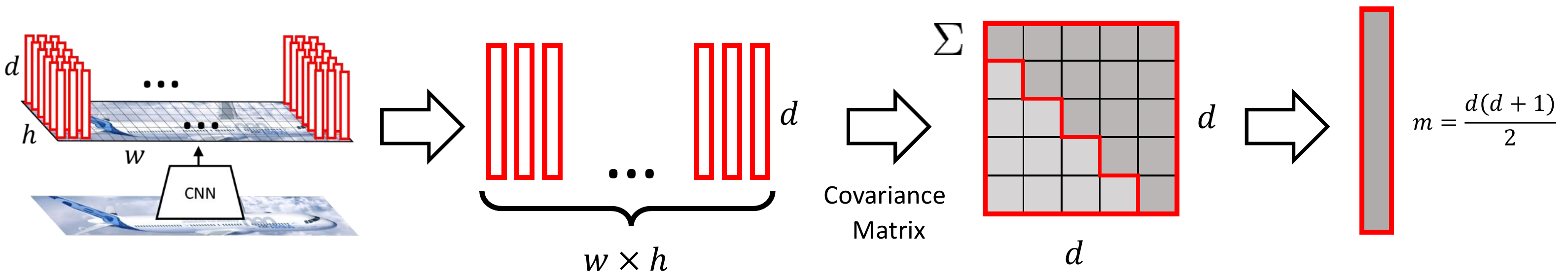}
    \caption{The $w \times h$ feature channels of dimension $d$ of the last convolutional layer of the CNN architecture are used to compute the covariance matrix. The $\frac{d(d+1)}{2}$-dimensional values of the upper triangular matrix constitute the internal feature representation vector that allows the model to determine the attention based long-range part-to-part relationships. The forward and backward propagation of the covariance in the adversarial optimization setting of Fig.~\ref{fig:architecture} are computed according to the iSQRT-COV approximated method.  }
   \Description{Illustration of covariance pooling on feature map: the covariance matrix is able to correlate different part of objects. }
    \label{fig:cov}
\end{figure*}

For labeled data, the general purpose-classification linear classifier already minimizes the distance between features and classifier prototypes. However, as in many other instance recognition tasks (i.e. face recognition, re-identification), features in ideal FGVC are expected to have smaller maximal intra-class distance than minimal inter-class distance under a suitably chosen metric space. The vanilla linear classifier cannot effectively satisfy this criterion \cite{liu2016large}. 
One simple method to enable convolutional neural networks to produce more discriminative features is imposing discriminative constraints on a hypersphere manifold by normalizing the vectors of the classifier weights \cite{liu2017sphereface}:

\begin{align}
\mathbf{w}^\prime_i = \frac{\mathbf{w}_i}{||\mathbf{w}_i||} \quad i=1,2, \dots K, \quad \mathbf{w}^\prime_i \in \mathbb{R}^{m} 
\label{eq:classifier}
\end{align}
where, as detailed in the next subsection,  $m=\frac{d(d+1)}{2}$ with $d$, is the channel dimension of the last convolutional layer of the architecture (Fig.~\ref{fig:cov}).

\subsubsection{Second-order Pooling Feature Extraction}
\label{subsec:Second-Order Pooling Feature Extraction}
Here we briefly recall the key elements for feature extraction as proposed in \cite{li2018towards}.
Let $\mathbf{X}\in\mathbb{R}^{w\times h\times d}$ be the output of the last convolutional layer of an architecture with spatial height $h$, width $w$ and channel $d$. The tensor is reshaped into a features matrix $\mathbf{X}$ consisting of $n=wh$ features of $d$ dimension.
Then second-order pooling is performed by computing the covariance matrix $\boldsymbol{\Sigma}=\mathbf{X}\bar{\mathbf{I}}\mathbf{X}^{T}$ as shown in Fig~\ref{fig:cov}. 
The covariance matrix $\boldsymbol{\Sigma}$ is a symmetric positive definite or semidefinite matrix, which can be factorized by EIG/SVD:
$
\boldsymbol{\Sigma} = \mathbf{U}\boldsymbol{\Lambda}\mathbf{U}^{T}
$.
Through EIG or SVD, the matrix power can be computed as:
$
\mathbf{Z}\stackrel{\vartriangle}{=}\boldsymbol{\Sigma}^{\alpha}=\mathbf{U}\mathrm{diag}(f(\lambda_{1}),\ldots,f(\lambda_{d}))\mathbf{U}^{T}
$
where the exponent $\alpha$ is a positive
real number. Empirically, the covariance matrix works best
when $\alpha$ is 0.5.

Matrix square root computation heavily depends on EIG or SVD operation; however, these methods don't have  a fast GPU implementation. To address this problem, \cite{li2018towards} proposes a fast method called iterative matrix square root normalization of covariance pooling (iSQRT-COV) that makes use of Newton-Schulz iteration. Let $\mathbf{Y}_{0}=\mathbf{A}$ and $\mathbf{Z}_{0}=\mathbf{I}$ for $k = 1,\dots, N$, the coupled iteration takes the following form:
\begin{align}
\mathbf{Y}_{k}&=\dfrac{1}{2}\mathbf{Y}_{k-1}(3\mathbf{I}-\mathbf{Z}_{k-1}\mathbf{Y}_{k-1})\nonumber \\
\mathbf{Z}_{k}&=\dfrac{1}{2}(3\mathbf{I}-\mathbf{Z}_{k-1}\mathbf{Y}_{k-1})\mathbf{Z}_{k-1}.
\label{eq:iteration}
\end{align}

These equations above can be computed with a single matrix product, which is more suitable for GPU implementation. 
In order to guarantee the convergence we pre-normalize covariance $\boldsymbol{\Sigma}$ by its trace or Frobenius norm, i.e.,
$
\mathbf{A}=\frac{1}{\mathrm{tr}(\boldsymbol{\Sigma})}\boldsymbol{\Sigma} \;\; \text{or}\;\;
\frac{1}{\|\boldsymbol{\Sigma}\|_{F}}\boldsymbol{\Sigma}.
$
Finally, in order to compensate the data magnitude caused by pre-normalization, the covariance matrix $\mathbf{Z}$ is calculated as: 

\begin{align}
\mathbf{Z}=\sqrt{\mathrm{tr}(\boldsymbol{\Sigma)}}\mathbf{Y}_{N}\;\; \text{or}\;\; \mathbf{Z}=\sqrt{\|\boldsymbol{\Sigma}\|_{F}}\mathbf{Y}_{N}.
\end{align}

The matrix $\mathbf{Z}$ is symmetric and has $\frac{d(d+1)}{2}$ different parameters that are gathered into a vector (i.e. the feature vector) which is the input of final classifier layer of Eq.~\ref{eq:classifier}.

\subsubsection{Handling Unlabeled Data with Adversarial Entropy Optimization}
\label{subsec:Handling Unlabeled Data with Adversarial Entropy Optimization}

The obtained features and the relative prototypes provide improved discriminative power and reduced low intra-class variation, respectively. Yet, the overall goal remains how to include a strategy to minimize the distance between the prototypes and the unlabeled examples. As this cannot be directly achieved with a linear classifier alone, we exploited the adversarial strategy originally proposed in \cite{ganin2015unsupervised,ganin2016domain}. The adversarial part of our strategy clusters features computed from unlabeled data around the classifier learned prototypes. 
Therefore, we train the feature extractor $F$ and the classifier $C$ to classify labeled and utilize standard cross-entropy minimization objective to extract discriminative features for the labeled data:
\begin{equation}
    \mathcal{L} = - \frac{1}{n_l}
\sum_{j=1}^{n_l}
\sum_{i=1}^{K}  \hat{y}_j\log p(y_j=i|\mathbf{x_j}),
\label{eq:celoss}
\end{equation}
where $ \hat{y}_j$ is the true class label for $x_j$.
The \textit{unlabeled data} are used to \textit{maximize} the entropy with respect to the classifier $C$ and to \textit{minimize} the entropy with respect the feature extractor $F$. The entropy is computed as follow:
\begin{equation}
H = -  
\frac{1}{n_u}
\sum_{j=1}^{n_u}
\sum_{i=1}^{K} p(y_j=i|\mathbf{x}_j) \log p(y_j=i|\mathbf{x}_j).
\label{eq:eloss}
\end{equation}
\begin{figure*}[t]
    \centering
    \includegraphics[width=0.6\textwidth]{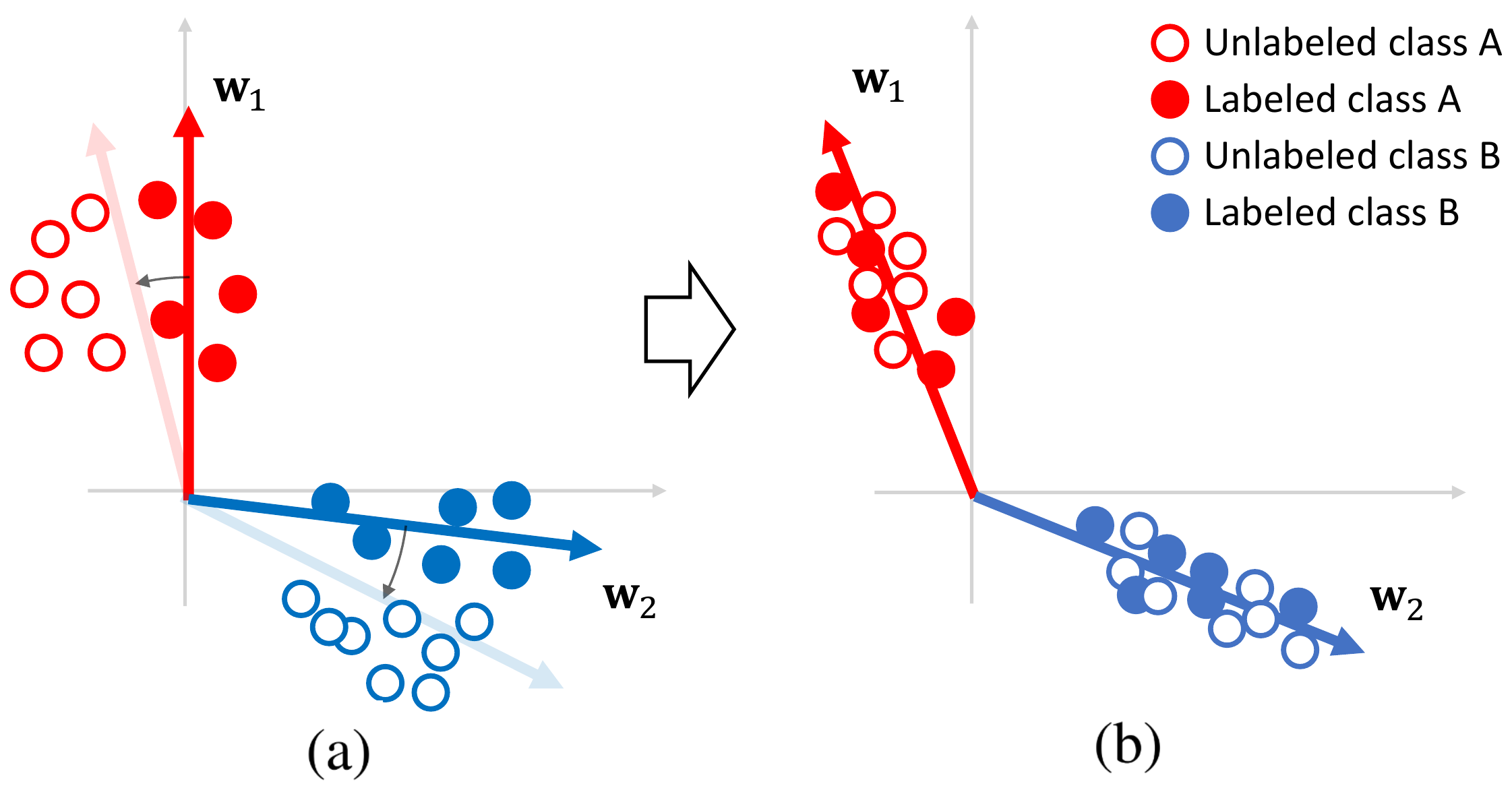}
    \caption{The adversarial learning intuition between the classifier $C$ and the feature extractor $F$ in a two class learning task. For visualization purposes, features are represented in a two-dimensional space. (a): High entropy between the classifier weight vectors $\mathbf{w}_1$ and $\mathbf{w}_2$ (i.e., the prototypes of the classifier $C$) and the unlabeled data features encourages the classifier weight vectors to ``move'' towards the unlabeled data features. (b): 
    This ``motion'' is counterbalanced by the low entropy which instead learns a representation (i.e. the feature extractor $F$) that tends to cluster labeled and unlabeled features around the estimated prototypes. }
    \label{fig:explanation}
\end{figure*}
The intuition that high entropy encourages the classifier weight vectors to ``move'' towards the unlabeled data features is illustrated in Fig.~\ref{fig:explanation}(a). 
This is due to the fact that high entropy, namely the maximization of Eq.~\ref{eq:eloss}, tends to achieve a uniform distribution of the softmax output probabilities that consequently encourage each prototype $\mathbf{w}_i$ to be similar to all the unlabeled features. 
This strategy can be considered as an \textit{``adversarial move''} of the classifier, whose intention is to \textit{``explore''} the representation space driven by the unlabeled data. 
This exploration is counterbalanced by the effect of low entropy that encourages learning a representation (i.e. the feature extractor $F$) that clusters labeled and unlabeled features around the estimated prototypes (Fig.~\ref{fig:explanation}(b)). Alternating these two opposite ``forces'' determines a sort of equilibrium in which the discriminative feature extractor $F$ and the classifier $C$ may have better explored the representation space as driven by unlabeled data. This can be considered as a form of soft pseudo-labeling strategy in which soft association can be eventually recovered.
This co-optimization can be formulated as an adversarial learning between $C$ and $F$ by weighted summing the two losses of Eq.~\ref{eq:celoss} and Eq.~\ref{eq:eloss} as follows: 
\begin{align}
\hat{\theta}_{F} &= \operatorname*{argmin}_{\theta_{F}} \mathcal{L} + \lambda H 
\label{eq:losses_H_min} 
\\
\hat{\theta}_{C} &= \operatorname*{argmin}_{\theta_{C}} \mathcal{L} - \lambda H,
\label{eq:losses_H_max}
\end{align}
where $\lambda { > 0} $ is the weighting factor between the two losses and $\theta_C$, $\theta_F$ are the learnable parameters of the classifier and the feature extractor, respectively.
$F$ and $C$ are co-optimized in two steps: in the first step, both $F$ and $C$ are optimized by minimizing the cross-entropy loss on labeled data. In the second step, $F$ and $C$ are optimized in opposite ways on unlabeled data, minimizing the entropy and maximizing the entropy, respectively (the signs of the entropy in the two equations of Eq.~\ref{eq:losses_H_min} and Eq.~\ref{eq:losses_H_max} are opposite). In particular, the negative sign of the entropy $H$ in Eq.~\ref{eq:losses_H_max} changes the
minimization problem into a maximization one. 
In the second step, input data follows the unlabeled path (Fig.~\ref{fig:architecture}), on which the classifier and the feature extractor are connected via a Gradient Reversal Layer (GRL) \cite{ganin2015unsupervised}. 
During forward propagation, GRL acts as an identity transform. During the back-propagation, GRL takes the gradient from the subsequent level, multiplies it by $-\lambda$ and passes it to the preceding layer (Fig.~\ref{fig:architecture}).
{By adding the gradient reversal layer, the  training process described above can be achieved through normal model training.}

\section{Experimental Results}
\subsection{FGVC Datasets}
\label{sec:Datasets}
Evaluation is performed on six datasets: \textit{FGVC Aircraft} \cite{maji2013fine}, \textit{Stanford Cars} \cite{krause20133d},  \textit{Cub-200-2011} \cite{wah2011caltech}, \textit{Stanford Dogs} \cite{KhoslaYaoJayadevaprakashFeiFei_FGVC2011}, Oxford Flowers \cite{Nilsback08} and the recently released SSL iNaturalist-Aves \error{\cite{su2021semisupervised}}.
\textit{FGVC Aircraft} 
is made up of 10,000 images of aircrafts, which are organized in four different hierarchies (Models, Variants, Families and Manufacturers). \textit{Stanford Cars} contains 16,185 images of 197 models of cars of various categories like SUVs, coupes, pickups, hatchbacks, station wagons. Each label reports manufacturer, model and production year. \textit{Cub-200-2011} is an extended version of CUB-200 dataset and it is composed of 11,788 images of 200 bird species. The labels of species have been gathered from Wikipedia and images have been retrieved from Flickr. Various metadata like bounding-boxes, attributes and annotations are available.
\textit{Oxford flower 102}  \cite{Nilsback08} is a 102 category dataset consisting of 102 flowers species. Images depict commonly occurring flowers in the United Kingdom. Each class consists of a variable number between 40 and 258 images. The training set consists of 2040 images while the test set consists of 6149 images. 
The \textit{Stanford Dogs Dataset} \cite{KhoslaYaoJayadevaprakashFeiFei_FGVC2011} contains 120 breeds of dogs from around the world. This dataset has been built using images and annotations from ImageNet for the task of fine-grained categorization. The training set is made up of  12000 images while test set consists of 8500 images. The SSL iNaturalist-Aves dataset is composed of images of 200 different species of birds split as in Tab.~\ref{tab:datasets}. We used the 3,959 labeled and the 26,640 unlabeled in-class images to train our model. The dataset also includes 122,208 unlabeled, out-of-class images we did not use. 

These datasets are rather complex due to: similarities existing among classes, occlusion, subtle intra-class variations, imbalance set. A summary of the evaluated datasets is shown in Tab.~\ref{tab:datasets}.

\begin{table*}[ht]
	\caption{A summary information for the evaluated FGVC datasets. Semi-Supervised iNaturalist-Aves   explicitly includes unlabeled data for classes outside the known ones (out-of-class). }
\centering
\small
	\begin{tabular}{cccccc}
	    \toprule
		{Dataset}       & {Classes \#} & {Train \#} & {Test \#} & {Unlabeled (In Class) \#} & Unlabeled (Out of Class) \# \\ \hline
		{FGVC-Aircraft \cite{maji2013fine}}             & 100                 & 6667                 & 3333  & -   &  -         \\ 
		{CUB-200-2011 \cite{wah2011caltech}}                  & 200                 & 5994                 & 5794 & -  &  -           \\
		{Stanford Cars \cite{krause20133d}}                  & 196                 & 8144                 & 8041   & -    &  -      \\
		{Oxford flowers \cite{Nilsback08}}                  & 102                 & 2040                 & 6149   & -    &  -      \\
		{Stanford Dogs \cite{KhoslaYaoJayadevaprakashFeiFei_FGVC2011}}                  & 120                 & 12000                 & 8580   & -     &  -     \\
		{Semi-Supervised iNaturalist-Aves \cite{su2021semisupervised}} & 200            & 4000   & 4000 & 26640 & 122208
		\\ \bottomrule
	\end{tabular}

	\label{tab:datasets}
\end{table*} 

\subsection{Baseline {methods}}
\label{sec:Baselines}
We compare the proposed method with the following baselines:
\begin{itemize}
    \item \textbf{Sup}: Supervised learning with standard linear classifier. No explicit methods for exploiting part-to-part relationships  are used {and unlabeled data is not considered}. 

    \item \textbf{Sup-Cov}: This is the second-order pooling (iSQRT-COV feature extractor) \cite{li2018towards}, trained in a \textit{fully supervised} setting with {only} cross-entropy loss minimization. 
    
    \item \textbf{Ent-Cov}: This is the second-order pooling (iSQRT-COV feature extractor) \cite{li2018towards}, trained using both labeled and unlabeled data minimizing both cross-entropy loss and  standard entropy minimization \cite{grandvalet2005semi}. Entropy is calculated on unlabeled data as defined in Eq.\ref{eq:eloss}.  In this case the adversarial optimization process is not considered. 
    
    \item \textbf{SSLFGC}: This is the SSL for Fine-Grained Classification (SSLFGC) method based on pseudo-label proposed in \cite{nartey2019semi}.  
\end{itemize}

\begin{figure}[t]
	\centering
	\includegraphics[width=0.5\linewidth]{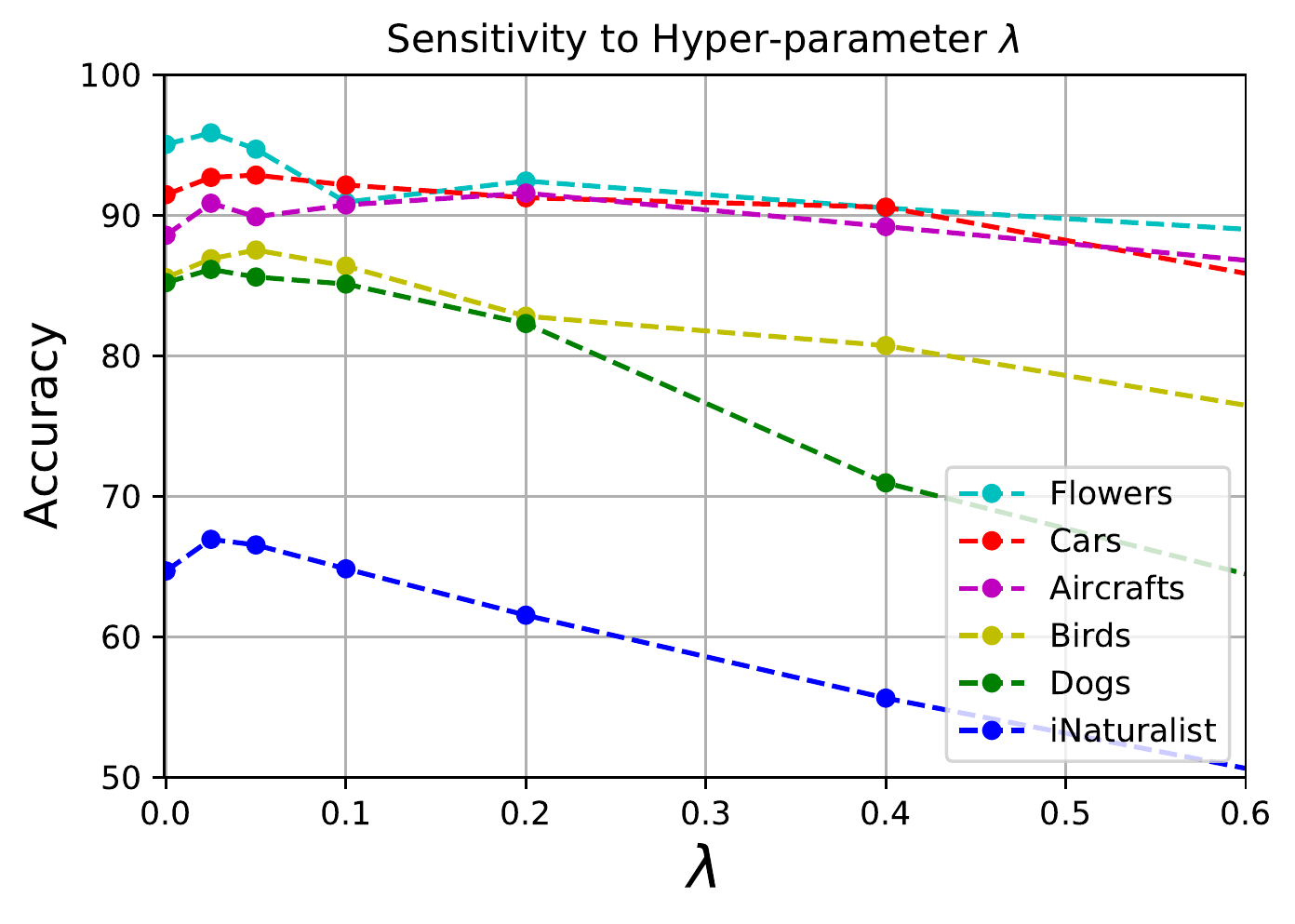}
	\caption{Accuracy as a function of the model parameter $\lambda$ evaluated for all the datasets. Values around $\lambda = 0.025$ achieve best accuracy for three of the datasets: \textit{Oxford Flowers}, \textit{Stanford Dogs} and \textit{SS iNaturalist-Aves}. \textit{FGVC Aircrafts} achieves the maximum accuracy around $\lambda=0.2$ while \textit{Stanford Dogs} and \textit{Stanford Cars} at $\lambda=0.05$.}
	 \Description{Illustration of accuracy variation depending on the lambda value. Thank to this study, we are able to determine the best value of lambda for each dataset.}
	\label{fig:lambda}
\end{figure}

\begin{table*}[t]
\caption{Comparison with other baselines on the Datasets reported in Tab.~\ref{tab:datasets}. We report the standard classification accuracy.
{ 
following the dataset splitting  described in section \ref{subsec:Evaluation on Standard Datasets}.}
Overall, our method largely outperforms other baselines for all evaluated datasets and for both architectures we used, with some exceptions as Flowers (where results are similar) and Dogs. 
}
\centering
\begin{tabular}{cccccccc}
\toprule
{Method}        & \multicolumn{6}{c}{{Dataset}}  & {\#Params } \\ \cmidrule{2-7}
\textbf{}       &  Aircrafts &  CUB-200-2011 &  Cars &  Flowers &  Dogs & SS iNat-Aves   & {}    \\  \midrule

{Sup }       & 87.38     & 85.50     & 88.35     & 94.05     & 74.49     & 52.7      & ResNet50 (25M)   \\ 
{Sup-Cov \cite{li2018towards}}       & 88.59     & 85.70     & 91.48     & 95.06     & 85.22     & 64.7      & ResNet50 (25M)   \\ 
{Ent-Cov}  {}       & 84.45     & 85.00     & 87.30     & 93.38     & 83.83     & 65.1      & ResNet50 (25M)   \\ 
\textbf{Ours} w/o Cov & 88.10     &  85.80     &  91.42     &  94.96     &  77.40     &  50.50      &  ResNet50 (25M) \\
{\bf Ours}      & 91.71     & 87.04     & 92.74     & 95.49     & 85.13     & 65.4      & ResNet50 (25M)   \\  \midrule
{SSLFGC \cite{nartey2019semi}}        & -         & 82.72     & 93.71     & 96.72 & \bf 91.67 & -         & Incept. ResNetV2 (56M)   \\  
{\bf Ours}      & \bf 91.74 & \bf 87.33 & \bf 93.83 & \bf 96.74     & 87.28     & \bf 69.85 & ResNet101 (44M)   \\

\bottomrule
\end{tabular}

\label{tab:comparison}
\end{table*}

\subsection{Evaluation on FGVC Datasets}
\label{subsec:Evaluation on Standard Datasets}
As one of the aim of these experiments is to compare our method with the semi-supervised method described in \cite{nartey2019semi}, we follow the datasets splits reported in the paper. According to this, the training set is partitioned into 75\% for training and 25\% for validation, while the entire test set is used as unlabeled data.
Our learning configuration includes two convolutional architectures of substantially different expressive power: the ResNet50 and the ResNet101 that have 25 and 44 million  parameters, respectively. Both architectures are pre-trained on the ILSVRC-2012 ImageNet dataset \cite{russakovsky2015imagenet}.
Images of any dataset are resized to 448x448. Data augmentation is performed according to basic image transformations as random crop, center crop and random horizontal flip. 
The proposed method is trained with SGD optimization and the learning rate is set to 0.0012 for both the CNN architectures and for the  second-order pooling (iSQRT-COV algorithm). The learning rate for the classifier is set to 0.003 following the implementation of \cite{li2018towards} and the batch size is set to 10. To balance efficiency and accuracy in the number of iterations $N$ in Eq.~\ref{eq:iteration}, we follow the same setting described in the original works \cite{li2018towards} \cite{BMVC2017_117}  and set $N$ to $5$ in all of our experiments.
Training is performed for 50k iterations and the validation in early stopping to prevent overfitting and for evaluating the sensitivity of the $\lambda$ hyper-parameter.

In particular, the sensitivity of the hyper-parameter $\lambda$ of Eq.~\ref{eq:losses_H_min} and  Eq.~\ref{eq:losses_H_max} is selected by grid-search in the interval $[0.025, 1]$. As shown in Fig.~\ref{fig:lambda} the resulted optimal values vary according to the specific dataset. Values around $\lambda = 0.025$ achieves the best accuracy for three of the datasets: \textit{Oxford Flowers}, \textit{Stanford Dogs} and \textit{SS iNaturalist-Aves}. \textit{FGVC Aircrafts} achieves the maximum accuracy around $\lambda=0.2$ while \textit{Stanford Dogs} and \textit{Stanford Cars} at $\lambda=0.05$. For the subsequent evaluations each dataset uses the value $\lambda$ which obtained the best accuracy.

As for $\lambda = 0$ the proposed method   optimizes the standard cross-entropy loss, it performs standard supervised learning. According to this, it is evident that the performance of the proposed method exploits favourably the unlabeled data in all the datasets when $\lambda > 0$. This because all the curves in Fig.~\ref{fig:lambda} increase in a neighbourhood of the origin. 

We compare with the baselines presented in Subsection~\ref{sec:Baselines} for all the datasets described in Subsection~\ref{sec:Datasets} and report all the results in Tab.~\ref{tab:comparison}. As it can be noticed, in comparison with Sup-Cov and Ent-Cov baselines, 
our approach is overall more effective, 
especially in \textit{FGVC Aircrafts}, \textit{CUB-200-2011} and \textit{Cars}. Comparable results are obtained in \textit{Flowers}. Conversely, SSLFGC obtains higher accuracy on \textit{Dogs} dataset by a clear margin. 
{The results of SSLFGC are taken from Tab.~5 of \cite{nartey2019semi} which reports the results considering the top-k most confident pseudo-labels images added to the training set. We choose their best results with $k=10$.} 
In this specific case SSLFGC is better than the current supervised learning state-of-the-art method \cite{zhuang2020learning} achieving an accuracy of 90.3\%. However, it is worth to notice that the SSLFGC approach exploits a CNN architecture with more parameters. 
  Tab.\ref{tab:comparison} also includes our approach without the Second-Order-Pooling, indicated as Ours w/o Cov, showing a clear detrimental effect on the overall performance. For the case of the iNaturalist-Aves dataset in which there is a large number of unlabeled data the improvement in exploiting the parts through the second order pooling is about 15\%.

Comparative feature evaluation according to t-SNE \cite{maaten2008visualizing} is shown in Fig.~\ref{fig:tsne} for the Sup-Cov supervised baseline and our SSL method for three of the six datasets. The figure shows that the method effectively leverages the structure of unsupervised data to obtain more discriminative features.

\subsection{Safety Evaluation}
\begin{figure}
    \subfigure[Ours]{
    \includegraphics[width=0.48\textwidth]{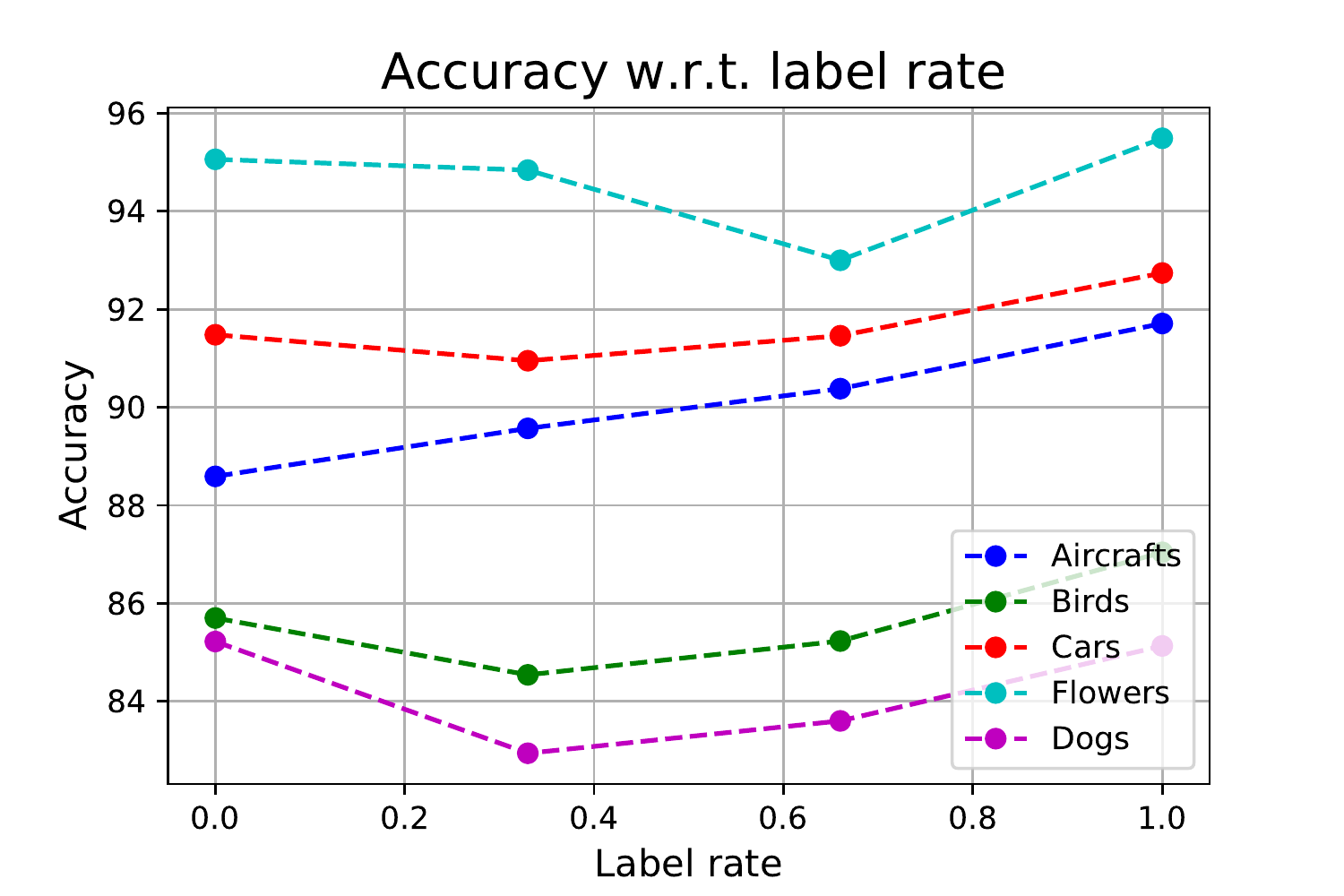}
    }
    \subfigure[Ent-Cov]{
    \includegraphics[width=0.48\textwidth]{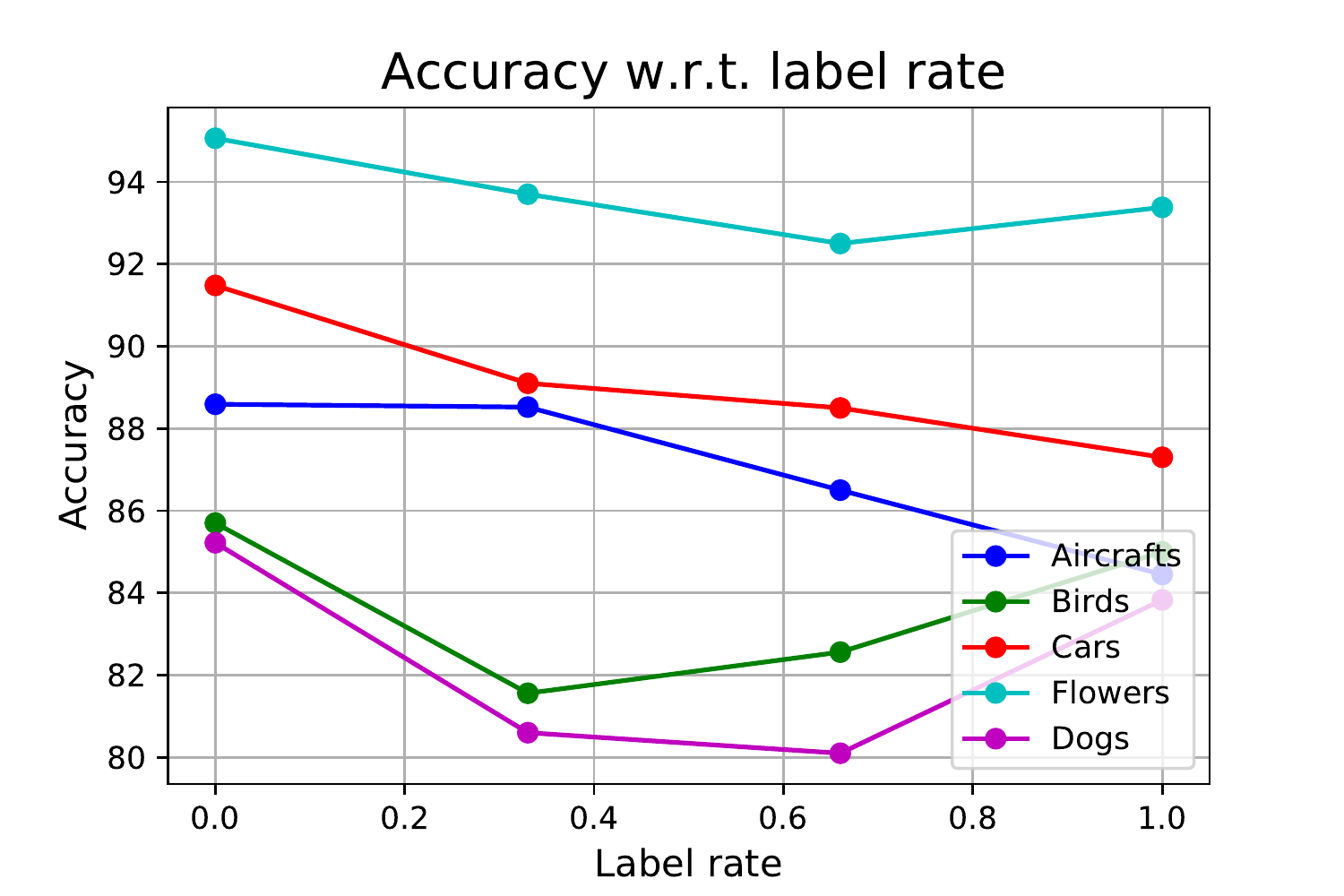}
    }
\caption{
Test accuracy for all FGVC datasets varying the rate number of unlabeled data (i.e. the label rate). The x-axis reports the label rate of test set used as unlabeled dataset. (a): Our method. (b): Ent-Cov. The proposed method improves performance as the number of the unlabeled data increases.
}
\label{fig:label_rate}
\end{figure}

\begin{figure}
    \hspace{-0.35cm}
    \subfigure[]{
    \includegraphics[width=0.35\columnwidth]{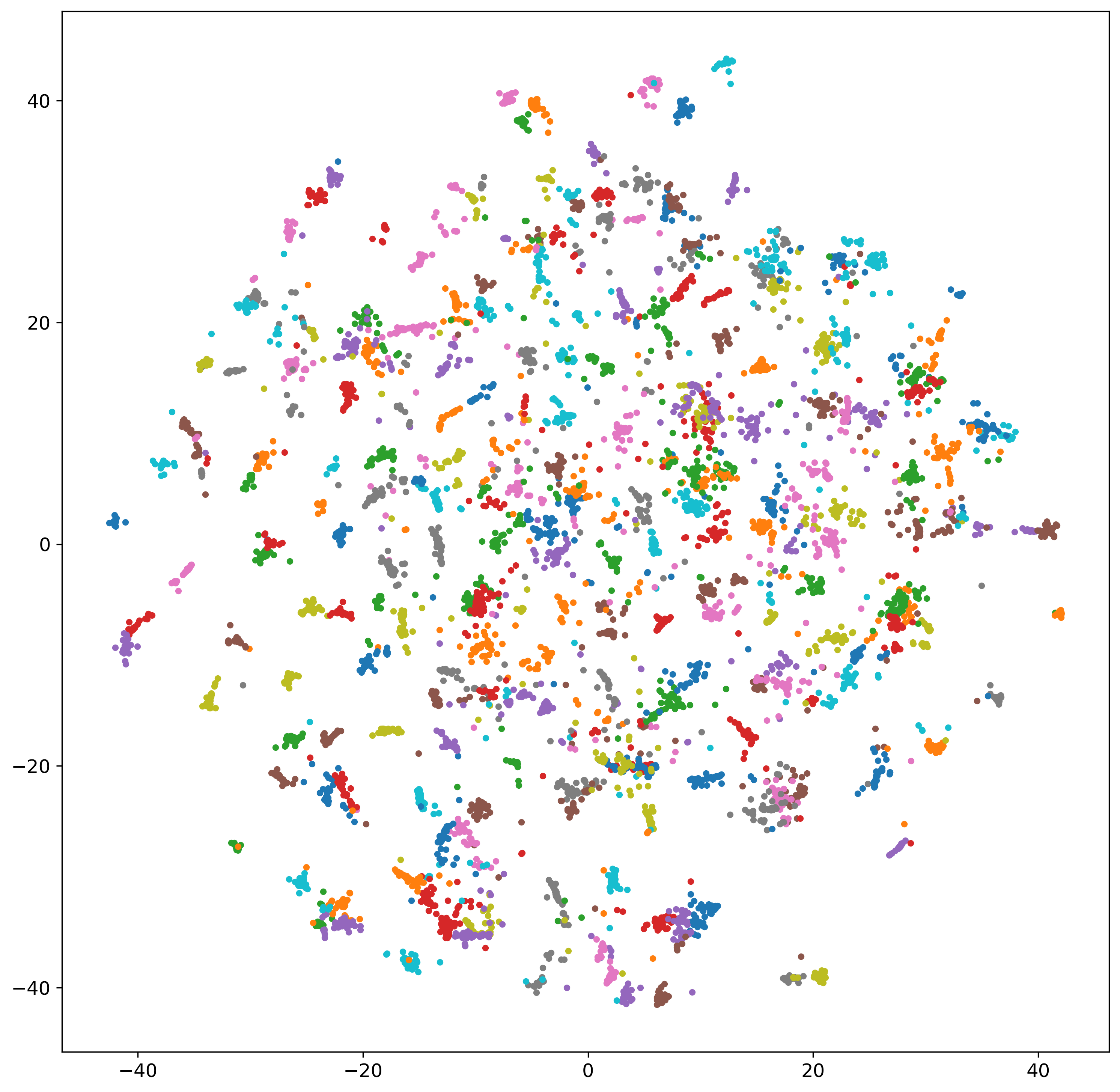}
    \includegraphics[width=0.35\columnwidth]{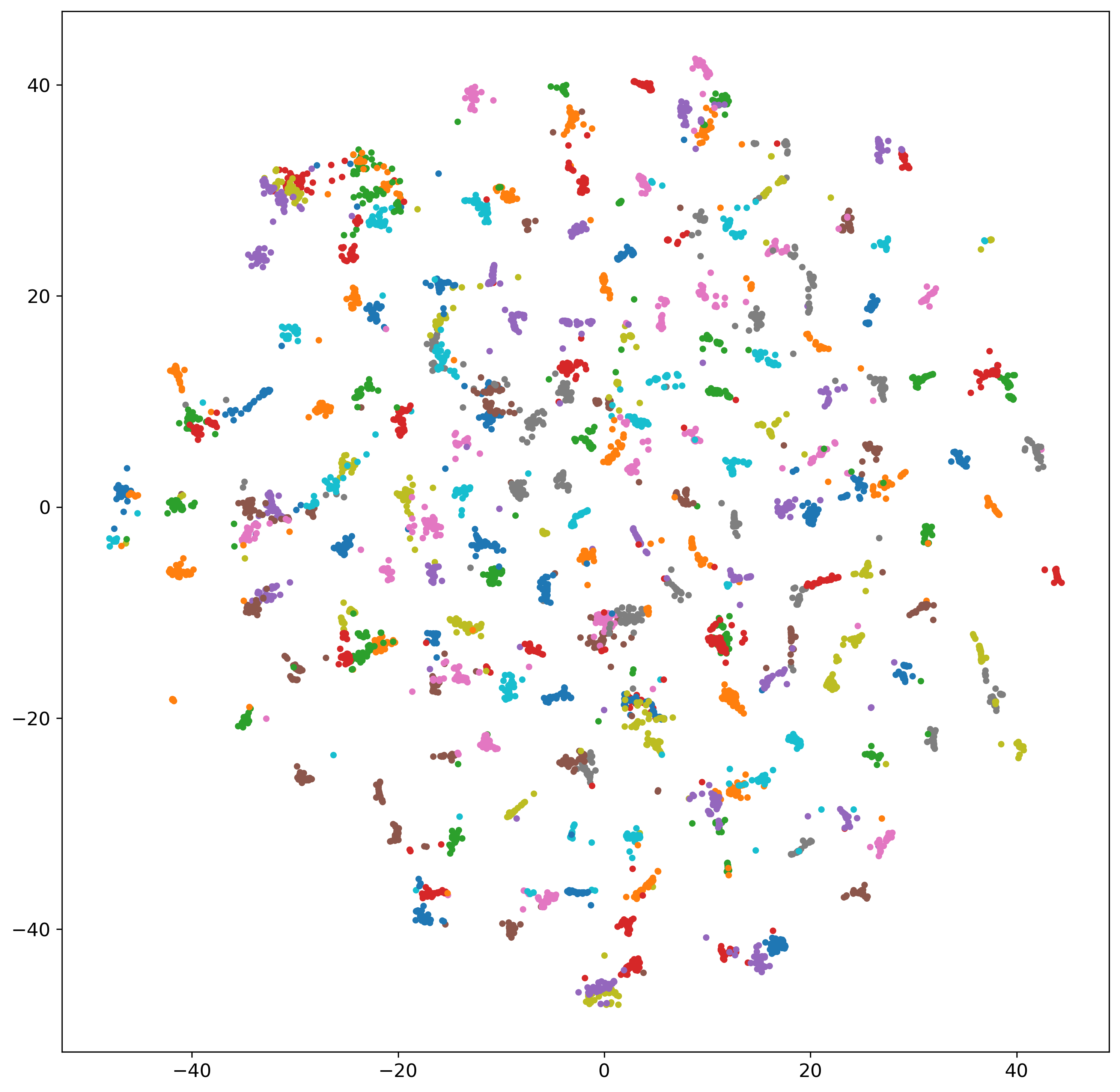}
    }
    \hspace{-0.35cm}
    \subfigure[]{
    \includegraphics[width=0.35\columnwidth]{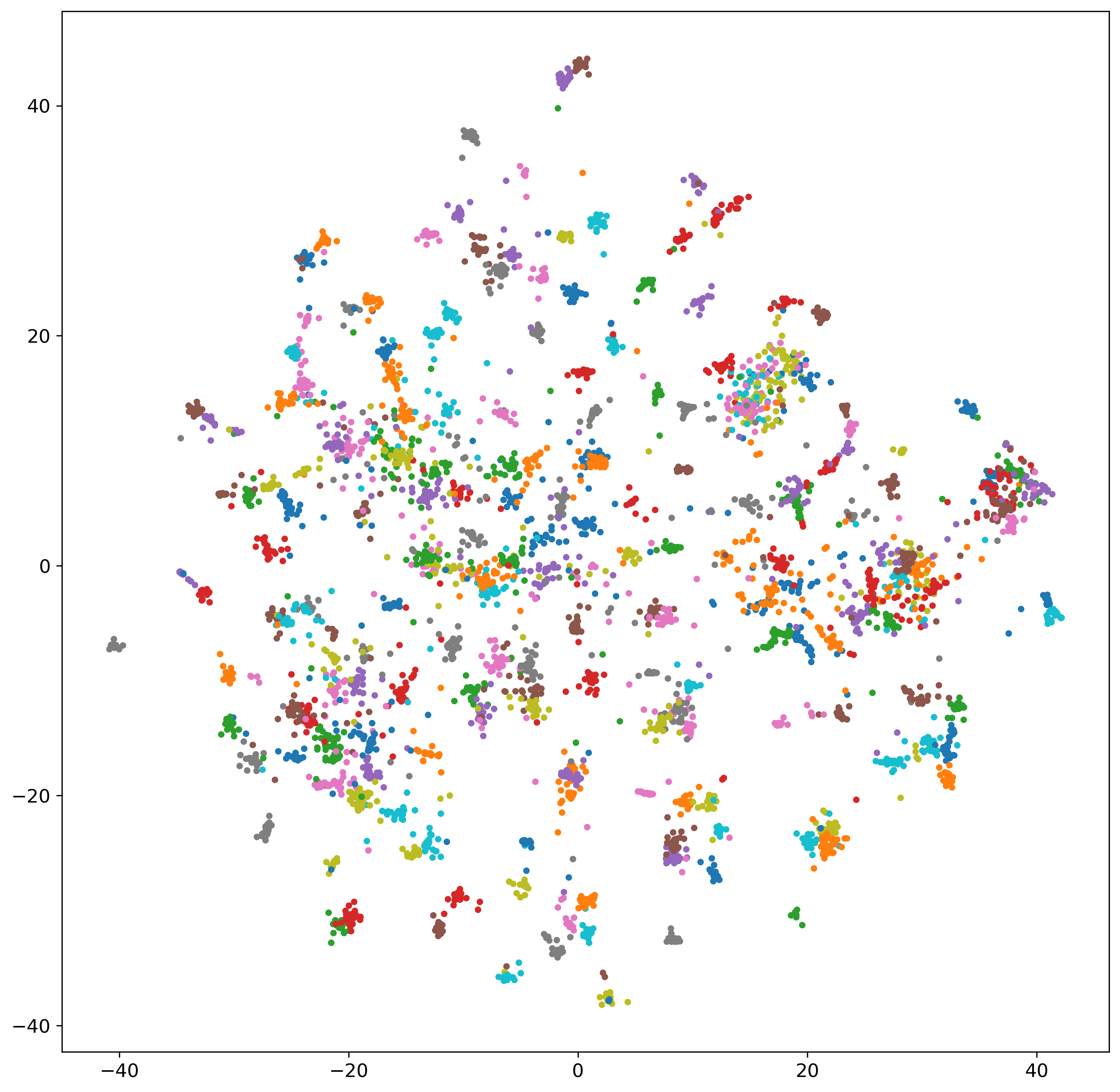}
    \includegraphics[width=0.35\columnwidth]{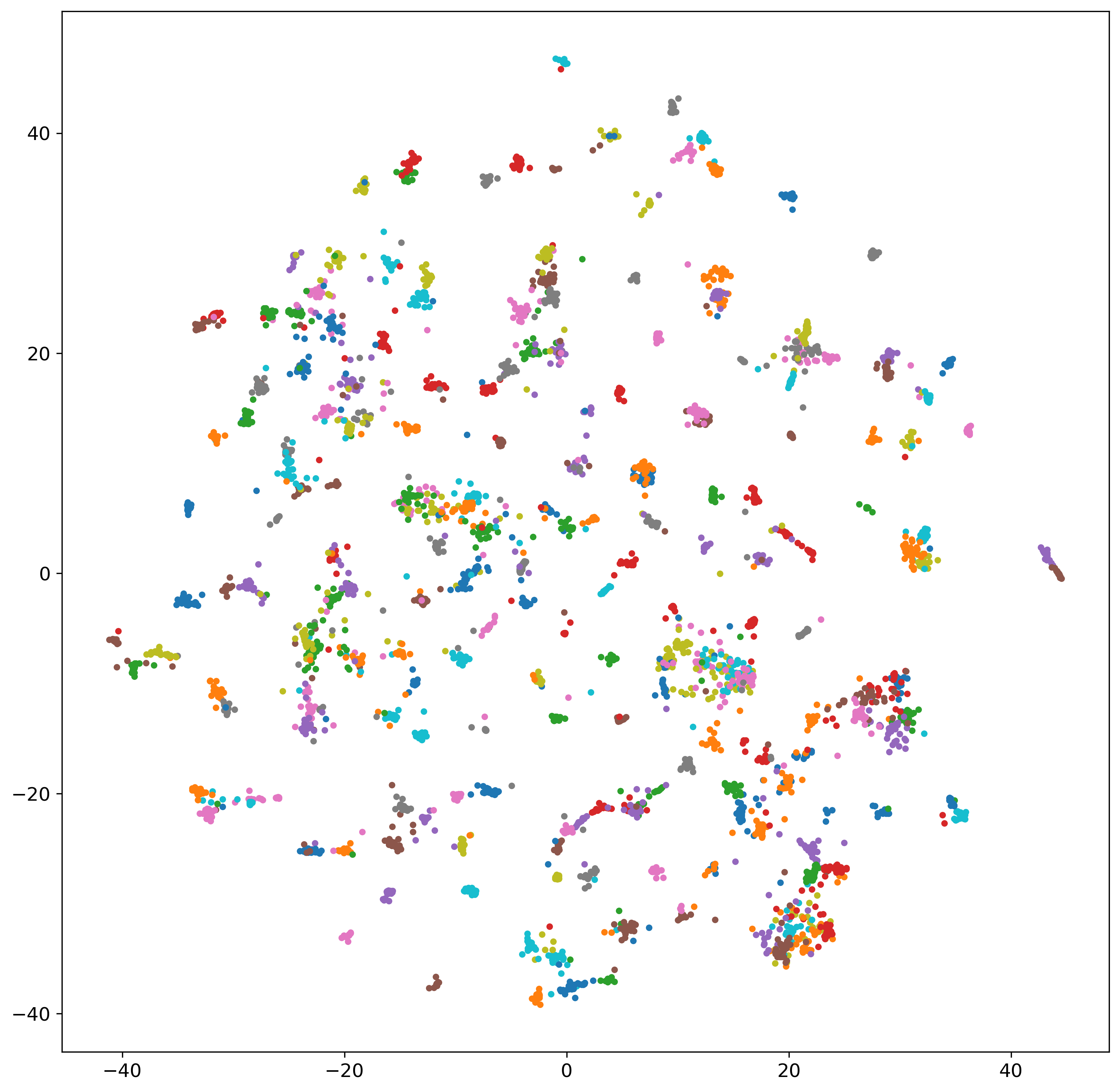}
     }
    \hspace{-0.35cm}
    \subfigure[]{
    \includegraphics[width=0.35\columnwidth]{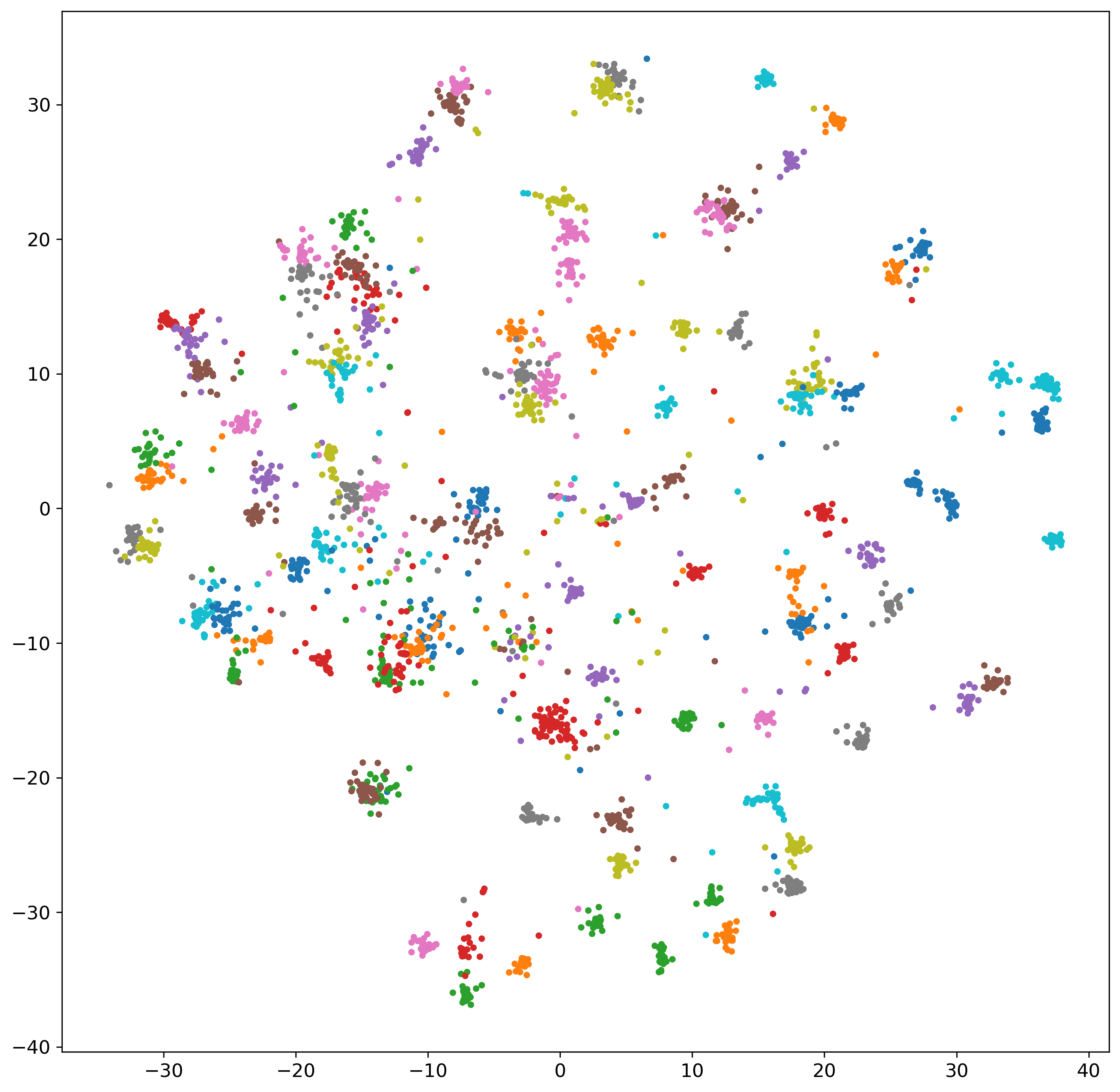}
    \includegraphics[width=0.35\columnwidth]{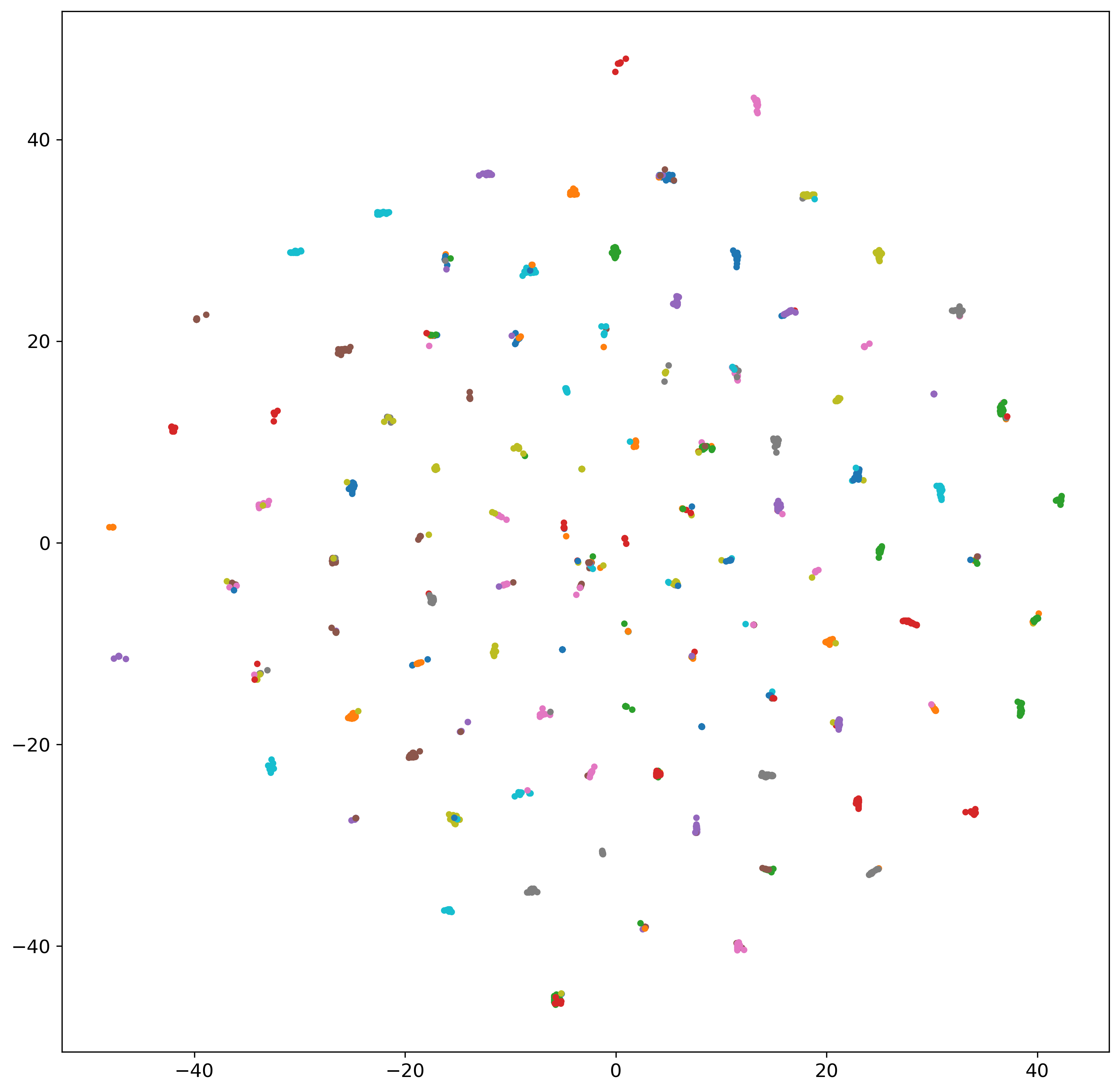}
    }
    \caption{Feature visualization with t-SNE.
    Qualitative comparison between features learned by the Sup-Cov baseline (\textit{left} column) and the method proposed in this paper (\textit{right} column) for the: \textit{(a)} Cars, \textit{(b)} Birds and \textit{(c)} Aircrafts datasets. As evidenced by the figures, the features learned by the proposed method are more discriminative.
    }
    \Description{Illustration of t-sne embedding: embeddings from semi-supervised are more discriminative and compact than the one from fully supervised learning }
    \label{fig:tsne}
\end{figure}

Semi-supervised classification methods could have lower performance than their supervised counterparts in some situations \cite{wang2013safety} and \cite{li2019safe}. 
It is therefore desirable to design secure semi-supervised classification methods which do not perform worse than their supervised counterparts. 
According to this, we evaluated the proposed method varying the amount of available unlabeled data using the same FGVC datasets split we used in the previous section. 
In this section we evaluate the \textit{label rate} (i.e., the amount of unlabeled data used for training). When the label rate $=0$, no unlabeled data is used for training (this is the case of the Sup-Cov baseline). When the label rate $= 1$,  the entire test set is used as unlabeled data.
As shown in Fig.~\ref{fig:label_rate}(a) the classification accuracy in all FGVC datasets increases as the label rate increases. 
Although performance degrades slightly at the beginning, on average the trend is that of positive growth. The proposed approach performs better at label rate 1 than at label rate 0 for all the evaluated datasets.
This result is apparently obvious but yet not to be taken for granted in SSL \cite{wang2013safety}.
Further possible causes of the slight degradation (i.e. not strict growth) at the beginning may also be found in the different levels of sensitivity of the algorithms while varying datasets \citep{oliver2018realistic}. To this end, we further evaluated in Fig.5(b) the behavior of our method in the absence of the adversarial learning strategy (i.e. \citep{grandvalet2005semi} with the second order pooling that we called Ent-Cov). As evidenced by the figure, the adversarial strategy significantly improves the overall capability of dealing with unlabeled data. For Aircrafts and Cars, for example, Ent-Cov is not able to take advantage of the unlabeled data while for the other datasets our method significantly improves the performance.
Contrarily, the Ent-Cov method doesn't efficiently leverage unlabeled data; the accuracy test for Flowers, Cars, Aircraft accuracy decreases as the number of unlabeled data grows up. For Dogs and Birds datasets, the trend of accuracy is similar to our method.

\subsection{Evaluation on the Semi-Supervised iNaturalist-Aves Dataset}

This dataset is significantly harder to evaluate using
the protocol we followed to compare with \cite{nartey2019semi}. This is due to the fact that the ratio between the number of labeled images and the unlabeled ones, except for \textit{Oxford Flowers}, is greater than one.  
We compare with Sup-Cov \cite{li2018towards} as described in the previous section and with the six different SSL methods evaluated in \cite{su2021realistic}:  Pseudo-Labeling \citep{lee2013pseudo},   
Curriculum Pseudo-Labeling, \cite{cascantebonilla2020curriculum},
FixMatch \cite{sohn2020fixmatch},
Self-Training \cite{su2021realistic}, 
MoCo (Momentum Contrast) \cite{he2020momentum}
and MoCo + Self-Training \cite{ChenKSNH20}.
Specifically, the Self-training baseline initially trains a teacher model with only labeled data, then transfers the knowledge to a student model by distillation \cite{hinton2015distilling} using both labeled and unlabeled data.
The MoCo + Self-Training 
performs a self-supervised pre-training with MoCo then removes the final MLP layers and adds a classification layer that is trained with labeled data. 
The results of the comparison are shown in Tab.~\ref{tab:chall}. 
Our method shows state-of-the-art performance with respect to the baselines with an accuracy result of 69.85\% and 65.4\% with the ResNet101 and ResNet50 architecture, respectively. The gain in classification accuracy from FixMatch (i.e. the best performing algorithm) using the same CNN backbone is 8\% (from 57.4\% to 65.4\%). 
As evidenced from the table the increase in classification accuracy is mostly due to the second order pooling layer and secondly by the adversarial strategy. This also confirms the results on the Sec.~\ref{subsec:Evaluation on Standard Datasets}.

One of the participants to the FGVC7-competition covered their training details in \cite{cui2020semi}. As no implementation is available, a fair comparison with our method is difficult to be conducted. The method employs an ensemble of eight CNN architectures for a total of about 500M parameters learned with batch size of 128 elements and several sophisticated training tricks (for example high-resolution images and advanced data augmentation like CutMix \cite{yun2019cutmix}), which are difficult to reproduce and which considerably improve the performance of about 20\% points. ImageNet based pretrained models are further improved by clustering with unlabeled out-of-class data.

\begin{table}[H] 
\centering
\caption{Results on the Semi-Supervised iNaturalist-Aves Dataset (FGVC7 challenge). Our method achieves a significant improvement by leveraging unsupervised data.}
\small
\begin{tabular}{ccc}
\toprule
Method   & Accuracy &\#Params \\ \midrule
Pseudo-Label            \cite{lee2013pseudo}                    & 54.4  & ResNet50 (25M)    \\ 
Curriculum Pseudo-Label \cite{cascantebonilla2020curriculum}    & 53.4  & ResNet50 (25M)    \\ 
FixMatch                \cite{sohn2020fixmatch}                 & 57.4  & ResNet50 (25M)    \\ 
Self-Training           \cite{su2021realistic}                  & 55.5  & ResNet50 (25M)    \\
MoCo                    \cite{he2020momentum}                   & 55.5  & ResNet50 (25M)    \\ 
MoCo + Self-Training    \cite{ChenKSNH20}                       & 52.7  & ResNet50 (25M)    \\ 
Sup                     \cite{su2021realistic}                  & 52.7  & ResNet50 (25M)    \\ 
Sup-Cov                 \cite{li2018towards}                    & 64.7  & ResNet50 (25M)    \\
\midrule
\textbf{Ours} w/o Cov                                           & 50.5  & ResNet50 (25M)    \\
\textbf{Ours}                                                   & 65.4  & ResNet50 (25M)    \\
\textbf{Ours}                                                   & \textbf{69.85} & ResNet101 (44M)   \\
\bottomrule
\end{tabular}
\label{tab:chall}
\end{table}

\subsection{Ablation of Mini-batch Composition}

At every iteration two mini-batches are sampled: one from the labeled data and the other from the unlabeled data. The two mini-batches are used to evaluate the objective in Eq.~\ref{eq:losses_H_min} and Eq.~\ref{eq:losses_H_max}, as shown in Fig. ~\ref{fig:architecture}, they follow two distinct backward path during back-propagation. We investigated the sensitivity of the relative and absolute size of the two mini-batches.Table ~\ref{tab:ablation} shows the effect of varying the relative composition of the two mini-batches on the accuracy evaluated on the six FGVC datasets.
As evidenced by the table, the configuration for the two mini-batches achieving the best average accuracy is with 10 labeled and 10 unlabeled examples (i.e. 86.25\%). Keeping the amount of labeled examples constant while increasing and decreasing of 5 examples the unlabeled mini-batch does not improve the average accuracy over the datasets. In both cases we obtained similar values of the average accuracy: 85.62 and 85.87.
As shown in the same table, increasing the size to 20 elements for each mini-batch does not improve the accuracy. 
This result suggests that there exists an inherent trade-off between the information carried by the two different mini-batches.

 \begin{table*}[ht]
  \caption{Sensitivity of the mini-batch composition. Best performance is achieved when:  1) the mini-batches for the two distinct paths contain the same number of examples and 2) when their size is 10.} 
 \centering
 \begin{tabular}{@{\extracolsep{4pt}}ccccccccccc}
 \toprule   
\#labeled/\#unlabeled & \multicolumn{6}{c}{Dataset} & Avg &\multicolumn{1}{c}{\#Params} \\
  \cmidrule{2-7} 
   & Aircrafts & CUB & Cars & Flowers & Dogs & SS iNat-Aves          \\ 
\midrule
     10/5  & 90.72 & 86.92 & 92.38 & 93.99 & 84.96 & 64.80 & 85.62 & ResNet50 (25M)   \\
     10/10 & 91.71 & 87.04 & 92.74 & 95.49 & 85.13 & 65.40 & \textbf{86.25} & ResNet50 (25M)   \\
     10/15 & 91.26 & 86.92 & 92.56 & 92.96 & 85.68 & 65.85 & 85.87 & ResNet50 (25M)   \\
     \midrule
      20/20 & 90.75 & 86.47 & 91.95 & 94.00 & 86.38 & 66.75 & 86.05 & ResNet50 (25M)   \\
 \bottomrule
 \end{tabular}
\label{tab:ablation}

 \end{table*}

\section{Conclusion}
We firstly introduced a Semi-Supervised Learning (SSL) method which combines ideas and components from adversarial entropy optimization and second-order pooling. Our main goal was reducing the prohibitive annotation cost of Fine-Grained Visual Categorization (FGVC) according to the SSL setting, a {combination of} problems which has not been investigated in the past. Through extensive experiments on FGVC datasets, including the very recent \textit{Semi-Supervised iNaturalist-Aves}, we found that the proposed method exhibits significantly improved performance compared to some baselines we included and to the only previous work that examined this problem. In almost all FGVC datasets we evaluated, we obtained state-of-the-art performance. Moreover, we firstly introduced the Gradient Reversal Layer (GRL), one of the key elements of adversarial entropy optimization to the SSL setting. This extension and its validation were left as future work in the original study \cite{ganin2015unsupervised,ganin2016domain}.
\\[0.5cm]
{\textbf{Acknowledgment}}: 
This work was partially  supported by the Italian MIUR within PRIN 2017, Project Grant 20172BH297: I-MALL. 
This research was also partially supported by Leonardo  Finmeccanica  S.p.A.

\bibliographystyle{ACM-Reference-Format}
\bibliography{sample-base}

\end{document}